# HSI-CNN: A Novel Convolution Neural Network for Hyperspectral Image

Yanan Luo[*], Jie Zou, Chengfei Yao, Tao Li, Gang Bai[**]
College of Computer and Control Engineering
Nankai University
Tianjin, China
E-mail: * luoyanan@mail.nankai.edu.cn, ** baigang@nankai.edu.cn

*Abstract*—With the development of deep learning, the performance of hyperspectral image (HSI) classification has been greatly improved in recent years. The shortage of training samples has become a bottleneck for further improvement of performance. In this paper, we propose a novel convolutional neural network framework for the characteristics of hyperspectral image data, called HSI-CNN. Firstly, the spectral-spatial feature is extracted from a target pixel and its neighbors. Then, a number of one-dimensional feature maps, obtained by convolution operation on spectral-spatial features, are stacked into a two-dimensional matrix. Finally, the two-dimensional matrix considered as an image is fed into standard CNN. This is why we call it HSI-CNN. In addition, we also implements two depth network classification models, called HSI-CNN+XGBoost and HSI-CapsNet, in order to compare the performance of our framework. Experiments show that the performance of hyperspectral image classification is improved efficiently with HSI-CNN framework. We evaluate the model's performance using four popular HSI datasets, which are the Kennedy Space Center (KSC), Indian Pines (IP), Pavia University scene (PU) and Salinas scene (SA). As far as we concerned, HSI-CNN has got the state-of-art accuracy among all methods we have known on these datasets of 99.28%, 99.09%, 99.42%, 98.95% separately.

## I. INTRODUCTION

Hyperspectral images (HSI) [1] are captured by specialized remote sensor on the aircraft and collected from the spectral data reflected by the ground objects in a certain area of the earth. Different from the image data recording the change of the spatial characteristics, the main record is the change of the spectral characteristics of the one point called pixel, including the space in the region, i.e. the position and distribution information of the surface object; and the spectrum, that is the reflection intensity of each pixel in different wavelength bands, two types of data. Hyperspectral images therefore contain a wealth of information, even though different substances belonging to the same species do not diminish their resolution at all because of the different characteristics of their reflected spectral information [2]. There is a very wide range of thematic applications in modern society, such as Ecological science [3, 4], Geological science [5], Hydrological science [6], Precision agriculture [7, 8].

Hyperspectral image classification was mainly through some of the prior knowledge to obtain spatial information and spectral information for classification in the earlier research. Using mathematical morphology to extract characteristics include size, orientation and contrast of the spatial structures present in the image [9, 10]. [11] – [13] proposed sparse representation or patch-based sparse representation as spatial features. The features of data changed from the single-pixel into neighborhood would extract richer information as spectral features. Using spectral-spatial information [9, 14] for training and classify the HSI data with the conventional method of machine learning, such as k-nearest-neighbors (kNN) [15], support vector machines (SVMs) [16, 17], random forests (RFs) [18] and so on. However, these methods often require strong background knowledge of HSI, and the process of extracting features is more troublesome and easy to lose important features.

Deep learning becomes more and more attractive in different fields such as image classification [19, 20]. Convolution Neural Network (CNN) is one of the most common network framework. The greatest advantage of it is that features can be extracted from the hidden layer in the network without too much preprocessing of the data. The convolution kernel with weight-sharing is used to extract the distributed feature expression in the whole window and get more abstract and more expressive features through multiple convolution, pooling and full-connection processing at different levels. Inspired by these good results, many researchers began to study how to use the neural network to solve the hyperspectral classification problem [21] – [23]. The main idea is to use the pixel information as input directly and construct a suitable convolution neural network for training, finally get good results. Hu et al [24] proposed a 5 layer CNN, and the emergence of a 3D-cube data extraction method [25] not only use the frequency domain information as before, but also combined with spatial information. The accuracy of the traditional method has been greatly improved. However, due to the unavoidable problem of hyperspectral images: high dimensionality and small sample sizes [18], the structure of the network could not build too deep to avoid overfitting.

The contributions of this paper are as follows:

*1)* We propose a novel convolutional neural network framework (HSI-CNN) for hyperspectial image classificaiton. HSI-CNN is a good trade-off between the number of training samples and the complexity of the network, and overcome overfitting.

*2)* We implement HSI-CNN + XGBoost method, that is, the XGBoost is considered as a substitution of the output layer of HSI-CNN in order to prevent overfitting. The comparison of the results of the two experiments, produced by HSI-CNN and HSI-CNN+XGBoost, also shows that HSI-CNN framework is not overfitting.

*3)* We modify the CapsNet [27] structure, called HSI-CapsNet, to fit hyperspectral image classification. The comparison of the results of the two experiments, obtained by HSI-CNN and HIS-CapsNet, shows that HIS-CapsNet does not bring expected benefits.

We evaluate the HSI-CNN's performance using four popular HSI datasets, which are the Kennedy Space Center (KSC), Indian Pines (IP), Pavia University scene (PU) and Salinas scene (SA). As far as we concerned, HSI-CNN has got the state-of-the-art accuracy among all methods on these datasets of 99.28%, 99.09%, 99.42%, 98.95% separately.

The remainder of this paper is organized as follows. In Section 2, we introduce some background of our work. Then we present HSI-CNN framework in Section 3. Experiments on four datasets are given in Section 4. A conclusion is made in Section 5.

## II. RELATED WORK

This section we will give a brief introduction of the latest methods for HSI classification. And some strategies are used for reference in our model.

### A. Preprocessing of Hyperspectral image data

The most important characteristic of the HSI is the combination of imaging and spectral detection techniques, while imaging the spatial features of the target, dozens of or even hundreds of narrow bands are scattered for each spatial pixel for continuous spectral coverage. The data so formed can be visually described as "three-dimensional data blocks," as shown in Figure 1(a). Where *x* and *y* represent two-dimensional planar pixel information coordinate axes and the third dimension (λ-axis) is a wavelength information coordinate axis. The single pixel with spectrum bands is labeled as a category's samples for training.

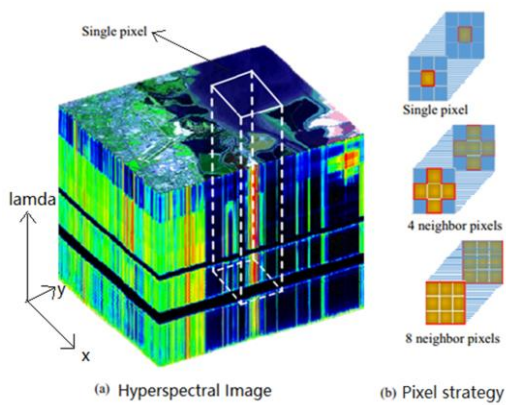

Fig. 1. The raw data of HSI. (a) The 3-dimensional data. (b) Cube data.

But in this way its only contains spectral information of HSI. The correlation between one pixel's neighborhood and itself is high, which have the same or similar characteristics. Leng et al [25] proposed an approach to extract a spectral cube of different spatial strategies, which separately mean using single-pixel, 4-neighbor pixels and 8-neighbor pixels. The pixel in the center is the one needed to be classified, as shown in Figure 1(b). Experimental results show that 8-neighbor pixels are the best, followed by 4-neighbor pixels and single pixel.

### B. CNN Based Hyperspectral image Classification

Hu et al [24] employed a CNN model to classify HSI directly in spectral domain, which can achieve better classification performance than some traditional methods. The structure of it is shown in figure 2. The network contains an input layer, a convolutional layer, a pooling layer, a fully connected layer and output layer. The convolution operation is similar as the operation of 2-dimension image, only its height is equal to 1.

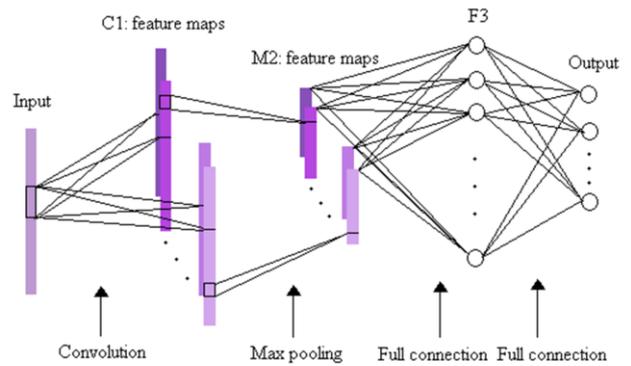

Fig. 2. The CNN structure proposed by Hu [24].

However, the above framework only takes the spectral characteristics into account and some spatial information will lost. Thus, cube-CNN [25] and 3D-CNN [28] are proposed to extract spectral-spatial features to improve the classification performance. How to organize the cube is mentioned in the previous sub section, there mainly point out the cube data or 3D data are convolved by 3D kernels as input to feed into net. The 3D data are similar to cube except the spatial scale, the latter is a patch (greater than $3 \times 3$) alternatively.

### C. CapsNet

Variant of CNN's have achieve state-of-the-art performance in many image relative tasks. However, there are two big shortcomings that CNN's cannot avoid: they can't take into account spatial hierarchies between features, they can't process rotational invariance. To address this defect, Hinton [27] propose a novel type of neural network named Capsule Network. By using the method of dynamic routing and vectoring result, the Capsule Network may take advantage of spatial hierarchies and keep rotation invariance. It gets impressive result on same datasets in [29], and we alter it to adapt the HSI data.

## III. HSI-CNN: A NOVEL CNN FOR HSI CLASSIFICATION

### A. HSI-CNN structure

We proposed a network structure that possesses novel processing strategy. Reshape 1-dimension array data to image-like 2-dimension matrix, which can make more completely use of the spectral and spatial information hidden in the original data. By referring to the idea of cube-CNN's selection of input data, 8-neighbor pixels labeled as the central pixel' categories are used as the input of our model. HSI-CNN proposed for HSI classification is shown in Figure.3. The network model is composed of 2 convolution layers, 1 reshape layer, 1 pooling layer and 3 fully connected layers where the last layer is softmax layer.

and softmax's nodes are equal to the number of categories of HSI data.

### B. Training procedure

There are 3 steps to train our model:

Step1: Extracting samples and dividing them into train and test datasets. The original data of each sample is extracted from the center of the pixel's 8 neighborhood cube data, and is labeled by the label of the central pixel.

Step2: Forward propagation. After a sample is fed into the HSI-CNN, the vectors obtained by the first layer of 3-dimensional convolution are reshaped into an image-like matrix. After the next 2-dimensional convolution and max pooling, the results are flattened into a vector, which will be

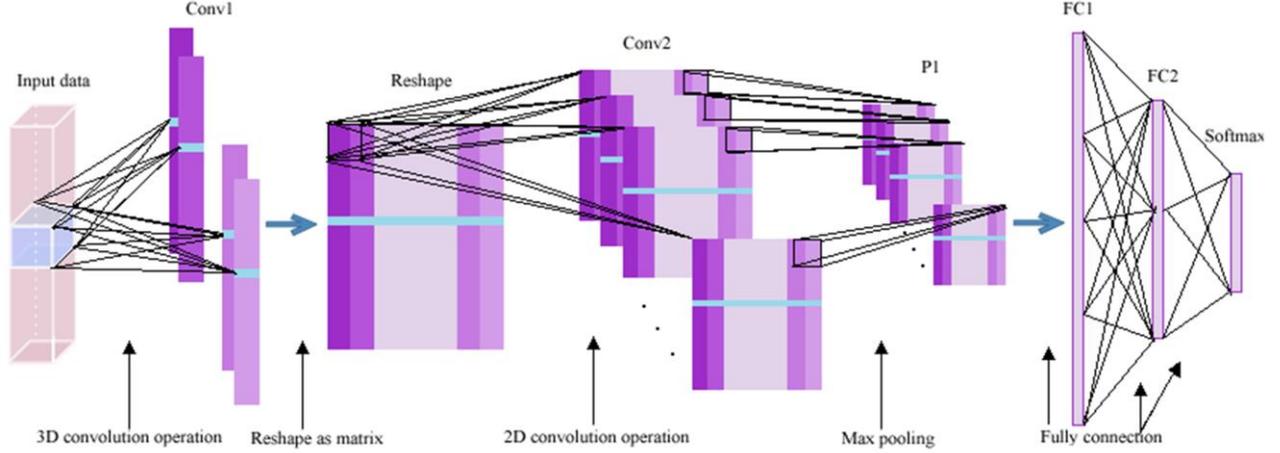

Fig. 3. The structure of HSI-CNN model.

The most important part of the model is the reshape layer. Since each feature vector is convoluted from the same original data with different convolution kernels which make these vectors have different representations of the particular features, so there is a strong correlation between these vectors. All the original vectors are directly stitched into a matrix whose height is constant and whose width changes from the original one to the number of feature vectors. After the reshape operation the data can be input as normal 2D image classification, so that in this way the scarce HSI data amount to a multiplied increase. While we put it into a depth network for training, we can also get good performance and avoid over-fitting.

More specifically, the 8-neighbor pixels cube is a $3 \times 3 \times n_{bands}$ data, where $n_{bands}$ is the number of spectral bands (or channels). The cube data are fed into the Conv1 convolved by $3 \times 3 \times n_{k1}$ kernels where the number of filters is $n_1$ and the stride is $s_1$. Therefore, Conv1 results are $n_1$ feature vectors each with height of $(n_{bands} - n_{k1}) / s_1$ and the width is 1. After reshape layer, the feature vectors becomes an image-like 2-dimension data of size $((n_{bands} - n_{k1}) / s_1) \times n_1$. The following operations include convolution, pooling and full connection is the same as the general deep learning network. Conv2 has 64 kernels size of $3 \times 3$, with stride $s_2$ the output becomes $h_1 \times n_2$, and $h_1 = ((n_{bands} - n_{k1}) / s_1 - 3) / s_2$, $n_2 = (n_1 - 3) / s_2$. After that, the 64 results are drawn into a vector as the input of the fully connected layer FC1 which has $n_3$ nodes. FC2 has $n_4$ nodes sent to fully connected layers. The last layer is calculated by *softmax* function, which indicates the probability of each class.

Step3: Updating weights. We use cross-entropy as a loss function to train the network. Each time we select a batch of data, optimizing it with a stochastic gradient descent (SGD) algorithm, and keep updating the parameters until convergence.

### C. HSI-CNN+XGBoost

HSI-CNN + XGBoost means that the XGBoost is considered as a substitution of the softmax layer of HSI-CNN in order to prevent overfitting. The idea has been proved that the similar method can gain significantly accuracy improvement in handwritten digits recognition applications [30].

### D. HSI-CapsNet

We start with a baseline model, Hinton's MNIST model, modify it by experiments for our HSI-CapsNet. The architecture is still with two convolutional layer and one fully connected layer. We use 3 fully connected layers to reconstruct the inputs as the regularization method. The regularization part is calculate as the sum of squared differences between the outputs of the last fully connected layer and the input, and in order to make share it does not dominate the total loss, the regularization part is scale down by 0.005.

## IV. DATASETS AND EXPERIMENTS

### A. Datasets

Four datasets are measured by our method including the Kennedy Space Center (KSC), Indian Pines (IP), Pavia University scene (PU) and Salinas scene (SA). For all the data, we randomly select 80% of each categories in dataset as train dataset to ensure every class is able to include, the remains are as test dataset. Furthermore, all the data values are normalized to zero mean and one variance.

The Kennedy Space Center (KSC) in Florida was acquired by the NASA AVIRIS (Airborne Visible/Infrared Imaging Spectrometer) instrument on March 23, 1996. The KSC data, gathered from an altitude of approximately 20 km, 176 bands were used and have 13 classes for the analysis.

The Indian Pines (IP) was gathered by AVIRIS sensor over the Indian Pines test site in North-western Indiana and consists of 145 times145 pixels and 200 spectral reflectance bands. The ground truth available is designated into 16 classes and is not all mutually exclusive.

The Pavia University scene (PU) was collected by ROSIS sensor during a flight campaign over Pavia, northern Italy. The number of spectral bands is 103 and consists of 610 times 610 pixels. The ground truths differentiate 9 classes.

The last dataset Salinas scene (SA) was acquired by the 224-band AVIRIS sensor over Salinas Valley, California. We discarded the 20 water absorption bands for research. The area covered comprises 512 lines by 217 samples and ground truth contains 16 classes.

### B. Experiments setup

Our experimental platform is a PC equipped with an Intel Core i7, 16 GB memory and Nvidia GPU.

In the four datasets, the kernel height $n_{k1}$ is 24 and stride $s_1$, $s_2$ is 9 and 1 separately. The next convolutional layer's

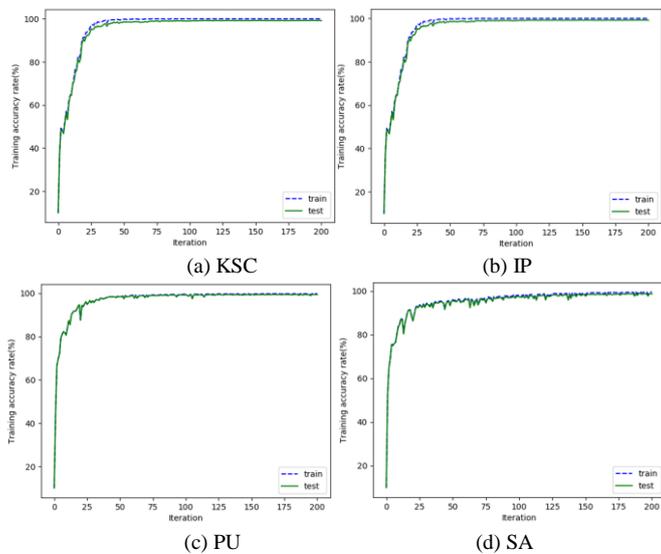

Fig.4 Accuracy of HSI-CNN on four datasets with iteration increasing, where the horizontal axis represents 300 iterations per number.

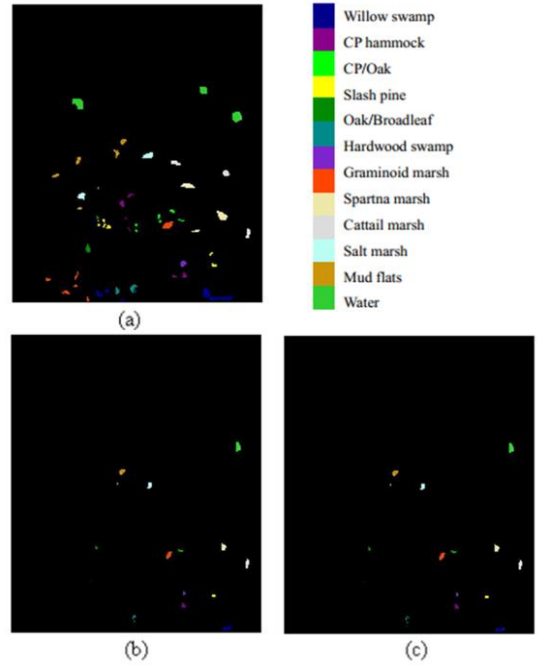

Fig. 5. The classification result of KSC with HSI-CNN. (a) The ground truth of KSC. (b) The ground truth of test dataset. (c) The HSI-CNN prediction. OA = 99.28%

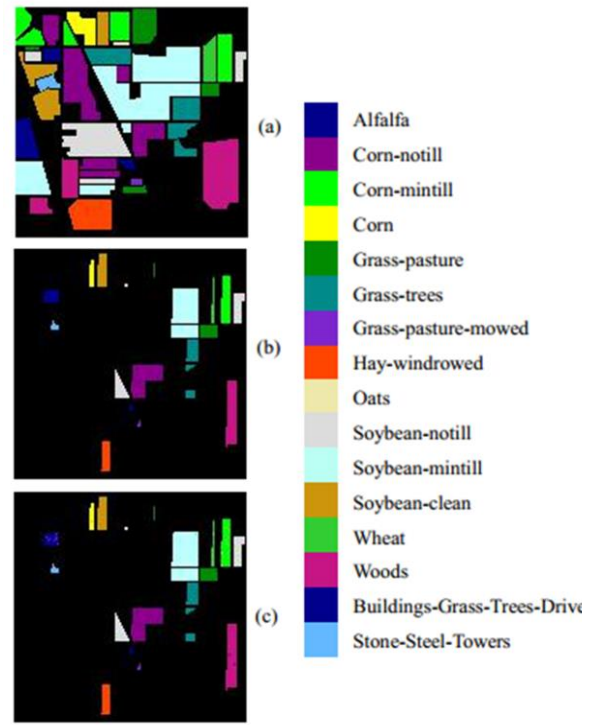

Fig. 6. The classification result of IP with HSI-CNN. (a) The ground truth of IP. (b) The ground truth of test dataset. (c) The HSI-CNN prediction. OA = 99.09%

kernel number 64 and the nodes of FC1 $n_3$ and FC2 $n_4$ is fixed with 1024 and 100 separately according to VGG and CNN. All the experiments are settled by the learning rate of

0.1, the decay term of 0.09 and batch size of 100. In the first experiment with KSC, the Conv1 kernels $n_1$ are 30. The Conv1 kernels $n_1$ are 60 on IP and SA datasets. And the PU dataset are configured with Conv1 kernels $n_1$ of 90.

We compared our method with two HSI classification approaches also proposed by ourselves: HSI-CapsNet and HSI-CNN+XGBoost. And one approaches proposed as CCS, the results of it are using derictly from [25].

For HSI-CapsNet, the first layer has 96, $1 \times 24$ convolutions kernels with a stride of 1 and sigmoid activation. The second layer is also a convolutional layer with 12 channels of convolutional 8D capsules. The kernel size and stride size is vary from data to data to make sure the second layer(PrimaryCapsulse layer) has [12, 5, 5] capsule output(each output is an 8D vector). The fianl layer has a 32D capusle per HSI pixel's class and each of this capusule recieve input from all the capsules in the layer below.

For HSI-CNN+XGBoost, we select the output of HSI-CNN's specific layer FC2 result to form the training features alternatively. Then use these new data to train XGBoost and classify. In our work, the XGBoost is considered as an additional classifier to evaluate the HSI-CNN extractor. Thus we implement it using the open source library xgboost of python module.

Overall accuracy (OA) is calculated to evaluate the performance of each model. Each cube we used is normalized to zero mean and unit variance. And we run the experiment 10 times for each datasets using each methods. The split ground truth datasets we use is fixed with the best performance of CNN for more convenient comparison.

*C. Results and Analyze*

The vary of accuracy with iteration increasing are shown in figure 4. Obviously four network begin to converge and stable after approximately $25 \times 300$ iterations. After convergence the network training accuracy are all closed to 100% and fitting well when testing. Thus, we can assert our structure HSI-CNN almost has learned the features hidden in original data without extra preposing. But due to the different amount of four datasets, the epochs that means how many forward passes the whole dataset among training are different. According to the batch size we choose, KSC, IP, PU and SA needs 42, 83, 323, 434 batches seperately in one epoch. It's clear that the more samples dataset has, the quicker network converges. Therefore, the greater the amout of data, the higher

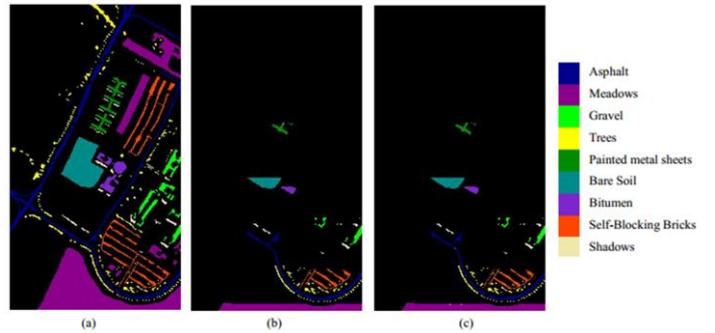

Fig. 7. The classification result of PU with our novel CNN. (a) The ground truth of PU. (b) The ground truth of test dataset. (c) The CNN prediction. OA = 99.52%

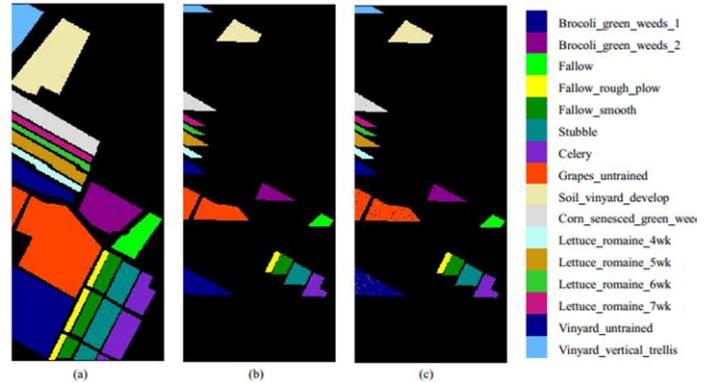

Fig. 8. The classification result of SA with our novel CNN. (a) The ground truth of SA. (b) The ground truth of test dataset. (c) The CNN prediction. OA = 98.95%

learning ability the network has.

Table 5 is the overall accuracy results of four dataset and provides the comparsion of performance with four classification methods. Firstly HSI-CNN gets the state-of-art performance around 99% compared with the latest method CCS and other two methods. And we can easily see that the HSI-CNN + XGBoost can almost gain the same performance as CCS. But the HSI-CapsNet are little weak for our datasets. Perhaps the model did not learn the relationship between different bands which are far apart in one pixel. More specific, HSI-CNN gained 0.5%, 1.28%, 0.1% and 3.91% higher classification accuracy. These results are also strongest prooves that our structure can learn and use the spectral-spatial information of HSI more effective. And the visible results of HSI-CNN are shown in Figure 5 – 8.

We also calculate the average accuracy of classification performance on four datasets. Among each dataset there are many classes that can gain accuracy more than 99% or even 100%. All the categories are classified correctly and get good accuracy more than 94%. This indicates the novel CNN we proposed are learning in balance even the samples of each categories are quite different.

TABLE I. CLASSIFICATION RESULTS OVERALL ACCURACY OF FOUR DATASETS

| Method<br>Datasets | CCS | HSI-CapsNet | HSI-CNN | HSI-CNN + XGBoost |
|---|---|---|---|---|
| KSC | 0.9871 | 0.9515 $\pm$ 0.0139 | **0.9928 $\pm$ 0.0040** | 0.9867 $\pm$ 0.0044 |
| IP | 0.9781 | 0.8949 $\pm$ 0.0073 | **0.9909 $\pm$ 0.0022** | 0.9857 $\pm$ 0.0044 |
| PU | 0.9945 | 0.9605 $\pm$ 0.0017 | **0.9952 $\pm$ 0.0015** | 0.9937 $\pm$ 0.0018 |
| SA | 0.9504 | 0.9575 $\pm$ 0.0063 | **0.9895 $\pm$ 0.0006** | 0.9891 $\pm$ 0.0006 |

## V. CONCLUSIONS

In this paper, we proposed a novel HSI classification model. The main idea is to reorganize data by using the correlation between convolution results and to splice the one-dimensional data into image-like two-dimensional data so as to deepen the network structure and enable the network to extract and distinguish the features better. Not only that, combined with the most popular machine learning model XGBoost as classifier while our HSI-CNN model extract the high level of feature. In addition, the application of hyperspectral data to Capsule network is proposed and implemented. In the end, our CNN model achieved the best overall accuracy, indicating that the ideas we proposed are feasible. We consider our method can provide an idea for all similar one-dimensional data analysis and research. The biggest challenge for our research is the lack of sample, which is also the core issue to be solved in the next step of our work.

ACKNOWLEDGMENT

This work is partially supported by the Natural Science Foundation of Tianjin (No. 16JCYBJC15200, No. 17JCQNJC00300), Tianjin Science and Technology Project (No. 15ZXDSGX00020), the National Key Research and Development Program of China (2016YFC0400709).

REFERENCES

[1] Landgrebe D. Hyperspectral image data analysis[J]. Signal Processing Magazine IEEE, 2002, 19(1):17-28.
[2] Pearlman J S, Barry P S, Segal C C, et al. Hyperion, a space-based imaging spectrometer[J]. Geoscience & Remote Sensing IEEE Transactions on, 2003, 41(6):1160-1173.
[3] Cochrane M A. Using vegetation reflectance variability for species level classification of hyperspectral data.[J]. International Journal of Remote Sensing, 2000, 21(10):2075-2087.
[4] Pontius J, Martin M, Plourde L, et al. Ash decline assessment in emerald ash borer-infested regions: A test of tree-level, hyperspectral technologies[J]. Remote Sensing of Environment, 2008, 112(5):2665-2676.
[5] Cloutis E A. Review Article Hyperspectral geological remote sensing: evaluation of analytical techniques[J]. International Journal of Remote Sensing, 1996, 17(12):2215-2242.
[6] Bulcock H H, Jewitt G P W. Spatial mapping of leaf area index using hyperspectral remote sensing for hydrological applications with a particular focus on canopy interception.[J]. Hydrology & Earth System Sciences, 2010, 14(2):383-392.
[7] Erives H, Fitzgerald G J. Automated registration of hyperspectral images for precision agriculture[J]. Computers & Electronics in Agriculture, 2005, 47(2):103-119.
[8] Lanthier Y, Bannari A, Haboudane D, et al. Hyperspectral Data Segmentation and Classification in Precision Agriculture: A Multi-Scale Analysis[C]// Geoscience and Remote Sensing Symposium, 2008. IGARSS 2008. IEEE International. IEEE, 2009:II-585 - II-588.
[9] Fauvel M, Tarabalka Y, Benediktsson J A, et al. Advances in Spectral-Spatial Classification of Hyperspectral Images[J]. Proceedings of the IEEE, 2013, 101(3):652-675.
[10] Serra J P. Image Analysis and Mathematical Morphology[J]. Biometrics, 1982, 39(2):536.
[11] Chen Y, Nasrabadi N M, Tran T D. Hyperspectral Image Classification Using Dictionary-Based Sparse Representation[J]. IEEE Transactions on Geoscience & Remote Sensing, 2011, 49(10):3973-3985.
[12] Chen Y, Nasrabadi N M, Tran T D. Hyperspectral Image Classification via Kernel Sparse Representation[J]. IEEE Transactions on Geoscience & Remote Sensing, 2012, 51(1):217-231.
[13] Yuan H. Robust patch-based sparse representation for hyperspectral image classification[J]. International Journal of Wavelets Multiresolution & Information Processing, 2017.
[14] Martin G, Plaza A. Spatial-Spectral Preprocessing Prior to Endmember Identification and Unmixing of Remotely Sensed Hyperspectral Data[J]. IEEE Journal of Selected Topics in Applied Earth Observations & Remote Sensing, 2012, 5(2):380-395.
[15] Ma L, Crawford M M, Tian J. Local Manifold Learning-Based $k$ -Nearest-Neighbor for Hyperspectral Image Classification[J]. IEEE Transactions on Geoscience & Remote Sensing, 2010, 48(11):4099-4109.
[16] Melgani F, Bruzzone L. Classification of hyperspectral remote sensing images with support vector machines[J]. IEEE Transactions on Geoscience & Remote Sensing, 2004, 42(8):1778-1790.]
[17] Ratle F, Camps-Valls G, Weston J. Semisupervised Neural Networks for Efficient Hyperspectral Image Classification[J]. IEEE Transactions on Geoscience & Remote Sensing, 2010, 48(5):2271-2282.
[18] Ham J, Chen Y, Crawford M M, et al. Investigation of the random forest framework for classification of hyperspectral data[J]. IEEE Transactions on Geoscience & Remote Sensing, 2005, 43(3):492-501.
[19] Lecun Y, Bottou L, Bengio Y, et al. Gradient-based learning applied to document recognition[J]. Proceedings of the IEEE, 1998, 86(11):2278-2324.
[20] Simonyan K, Zisserman A. Very Deep Convolutional Networks for Large-Scale Image Recognition[J]. Computer Science, 2014.
[21] Chen Y, Lin Z, Zhao X, et al. Deep Learning-Based Classification of Hyperspectral Data[J]. IEEE Journal of Selected Topics in Applied Earth Observations & Remote Sensing, 2017, 7(6):2094-2107.
[22] Zhang H, Li Y, Zhang Y, et al. Spectral-spatial classification of hyperspectral imagery using a dual-channel convolutional neural network[J]. Remote Sensing Letters, 2017, 8(5):438-447.
[23] Yu S, Jia S, Xu C. Convolutional neural networks for hyperspectral image classification[J]. Neurocomputing, 2016, 219.
[24] Hu W, Huang Y, Wei L, et al. Deep Convolutional Neural Networks for Hyperspectral Image Classification[J]. Journal of Sensors, 2015, 2015(2):1-12.
[25] Leng J, Li T, Bai G, et al. Cube-CNN-SVM: A Novel Hyperspectral Image Classification Method[C]// IEEE, International Conference on TOOLS with Artificial Intelligence. IEEE, 2017:1027-1034.
[26] Chen T, Guestrin C. XGBoost: A Scalable Tree Boosting System[J]. 2016:785-794.
[27] Sabour S, Frosst N, Hinton G E. Dynamic Routing Between Capsules[J]. 2017.
[28] Li Y, Zhang H, Shen Q. Spectral-Spatial Classification of Hyperspectral Imagery with 3D Convolutional Neural Network[J]. Remote Sensing, 2017, 9(1):67.
[29] Xi E, Bing S, Jin Y. Capsule Network Performance on Complex Data[J]. 2017.
[30] Niu X X, Suen C Y, "A novel hybrid CNN-SVM classifier for recognizing handwritten digits," Pattern Recognition, 2012, 45(4), pp.1318-1325.